\documentclass[runningheads]{llncs}
\usepackage{graphicx}
\usepackage{csquotes}
\begin{document}
\title{Interpretable Machine Learning -- A Brief History, State-of-the-Art and Challenges\thanks{This project is funded by the Bavarian State Ministry of Science and the Arts and coordinated by the Bavarian Research Institute for Digital Transformation (bidt) and supported by the German Federal Ministry of Education and Research (BMBF) under Grant No. 01IS18036A.
The authors of this work take full responsibilities for its content.
}}
\titlerunning{IML - History, Methods, Challenges}
%
\author{Christoph Molnar\inst{1}\orcidID{0000-0003-2331-868X}
\and
Giuseppe Casalicchio\inst{1}\orcidID{0000-0001-5324-5966}
\and
Bernd Bischl\inst{1}\orcidID{0000-0001-6002-6980}
}
\authorrunning{Molnar et al.}
%
\institute{Department of Statistics, LMU Munich\\Ludwigstr. 33, 80539 Munich, Germany
\\
\email{christoph.molnar@stat.uni-muenchen.de}}
\maketitle              
\begin{abstract}
We present a brief history of the field of interpretable machine learning (IML), give an overview of state-of-the-art interpretation methods and discuss challenges.
Research in IML has boomed in recent years.
As young as the field is, it has over 200 years old roots in regression modeling and rule-based machine learning, starting in the 1960s.
Recently, many new IML methods have been proposed, many of them model-agnostic, but also interpretation techniques specific to deep learning and tree-based ensembles.
IML methods either directly analyze model components, study sensitivity to input perturbations, or analyze local or global surrogate approximations of the ML model.
The field approaches a state of readiness and stability, with many methods not only proposed in research, but also implemented in open-source software.
But many important challenges remain for IML, such as dealing with dependent features, causal interpretation, and uncertainty estimation, which need to be resolved for its successful application to scientific problems.
A further challenge is a missing rigorous definition of interpretability, which is accepted by the community.
To address the challenges and advance the field, we urge to recall our roots of interpretable, data-driven modeling in statistics and (rule-based) ML, but also to consider other areas such as sensitivity analysis, causal inference, and the social sciences.

\end{abstract}
\section{Introduction}

Interpretability is often a deciding factor when a machine learning (ML) model is used in a product, a decision process, or in research.
Interpretable machine learning (IML)\footnote{Sometimes the term Explainable AI is used.} methods can be used to discover knowledge, to debug or justify the model and its predictions, and to control and improve the model \cite{adadi2018peeking}.
In this paper, we take a look at the historical building blocks of IML and give an overview of methods to interpret models.
We argue that IML has reached a state of readiness, but some challenges remain.

\section{A Brief History of IML}
A lot of IML research happened in the last couple of years.
But learning interpretable models from data has a much longer tradition.
Linear regression models were used by Gauss, Legendre, and Quetelet \cite{stigler1986history,legendre1805nouvelles,gauss1809theoria,quetelet1827recherches} as early as the beginning of the 19th century and have since then grown into a vast array of regression analysis tools \cite{tibshirani1996regression,santosa1986linear}, for example, generalized additive models \cite{hastie1990generalized} and elastic net \cite{zou2005regularization}.
The philosophy behind these statistical models is usually to make certain distributional assumptions or to restrict the model complexity beforehand and thereby imposing intrinsic interpretability of the model.

In ML, a slightly different modeling approach is pursued.
Instead of restricting the model complexity beforehand, ML algorithms usually follow a non-linear, non-parametric approach, where
model complexity is controlled through one or more hyperparameters and selected via cross-validation.
This flexibility often results in less interpretable models with good predictive performance. 
A lot of ML research began in the second half of the 20th century with research on, for example, support vector machines in 1974 \cite{vapnik1974theory}, early important work on neural networks in the 1960s \cite{schmidhuber2015deep}, and boosting in 1990 \cite{schapire1990strength}.
Rule-based ML, which covers decision rules and decision trees, has been an active research area since the middle of the 20th century \cite{furnkranz2012foundations}.

\begin{figure}
  \includegraphics[width=\textwidth]{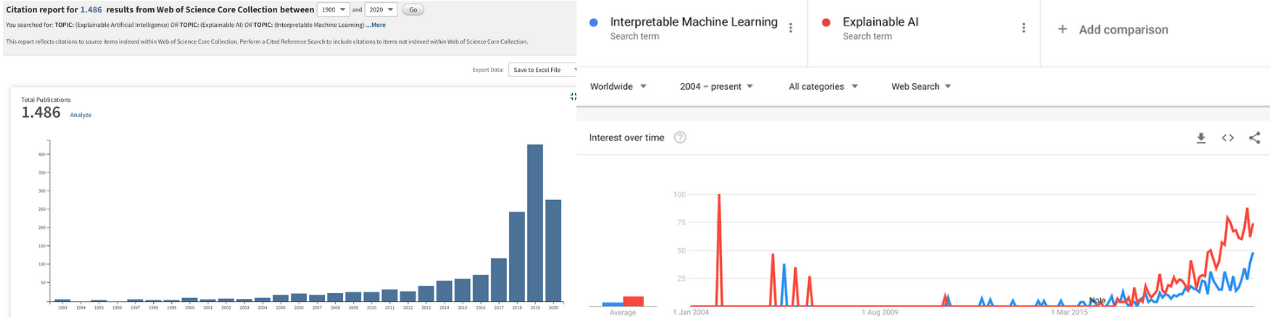}
  \caption{Left: Citation count for research articles with keywords \enquote{Interpretable Machine Learning} or \enquote{Explainable AI} on Web of Science (accessed August 10, 2020). Right: Google search trends for \enquote{Interpretable Machine Learning} and \enquote{Explainable AI} (accessed August 10, 2020).}
  \label{fig:count}
\end{figure}

While ML algorithms usually focus on predictive performance, work on interpretability in ML -- although underexplored -- has existed for many years. 
The built-in feature importance measure of random forests \cite{breiman2001random} was one of the important IML milestones.\footnote{The random forest paper has been cited  over 60,000 times (Google Scholar; September 2020) and there are many papers improving the importance measure (\cite{strobl2008conditional,strobl2007bias,hapfelmeier2014new,ishwaran2007variable}) which are also cited frequently.}
In the 2010s came the deep learning hype, after a deep neural network won the ImageNet challenge.
A few years after that, the IML field really took off (around 2015), judging by the frequency of the search terms "Interpretable Machine Learning" and "Explainable AI" on Google (Figure \ref{fig:count}, right) and papers published with these terms (Figure \ref{fig:count}, left).
Since then, many model-agnostic explanation methods have been introduced, which work for different types of ML models. But also model-specific explanation methods have been developed, for example, to interpret deep neural networks or tree ensembles.
Regression analysis and rule-based ML remain important and active research areas to this day and are blending together (e.g., model-based trees \cite{zeileis2008model}, RuleFit \cite{friedman2008predictive}).
Many extensions of the linear regression model exist \cite{hastie1990generalized,fahrmeir2013multivariate,gelman2006data} and new extensions are proposed until today \cite{fasiolo2020scalable,caruana2015intelligible,fasiolo2020fast,ustun2016supersparse}. 
Rule-based ML also remains an active area of research (for example, \cite{wang2015falling,letham2015interpretable,hothorn2015ctree}).
Both regression models and rule-based ML serve as stand-alone ML algorithms, but also as building blocks for many IML approaches.

\section{Today}

IML has reached a first state of readiness.
Research-wise, the field is maturing in terms of methods surveys \cite{molnar2019,guidotti2018survey,vilone2020explainable,rosenfeld2019explainability,adadi2018peeking,anjomshoae2019explainable,du2019techniques,carvalho2019machine},
further consolidation of terms and knowledge \cite{hall2019systematic,doshi2017towards,murdoch2019definitions,samek2019towards,preece2018stakeholders,chromik2020taxonomy}, and work about defining interpretability or evaluation of IML methods \cite{mohseni2018multidisciplinary,mohseni2018human,ribeiro2016should,hoffman2018metrics}.
We have a better understanding of weaknesses of IML methods in general \cite{molnar2019,molnar2020pitfalls}, but also specifically for methods such as permutation feature importance \cite{hooker2019please,strobl2008conditional,apley2016visualizing,strobl2007bias}, Shapley values \cite{janzing2019feature,sundararajan2019many}, counterfactual explanations \cite{laugel2019dangers}, partial dependence plots \cite{hooker2019please,hooker2007generalized,apley2016visualizing} and saliency maps \cite{adebayo2018sanity}.
Open source software with implementations of various IML methods is available, for example, \textit{iml} \cite{iml} and \textit{DALEX} \cite{biecek2018dalex} for R \cite{Rlang} and \textit{Alibi} \cite{klaise2020alibi}  and \textit{InterpretML} \cite{nori2019interpretml} for Python.
Regulation such as GDPR and the need for ML trustability, transparency and fairness have sparked a discussion around further needs of interpretability \cite{wachter2017counterfactual}.
IML has also arrived in industry \cite{gade2019explainable}, there are startups that focus on ML interpretability and also big tech companies offer software \cite{wexler2019if,arya2020ai,hall2017machine}.

\section{IML Methods}

\begin{figure}
    \centering
    \begin{minipage}{0.45\textwidth}
        \centering
        \includegraphics[width=0.9\textwidth]{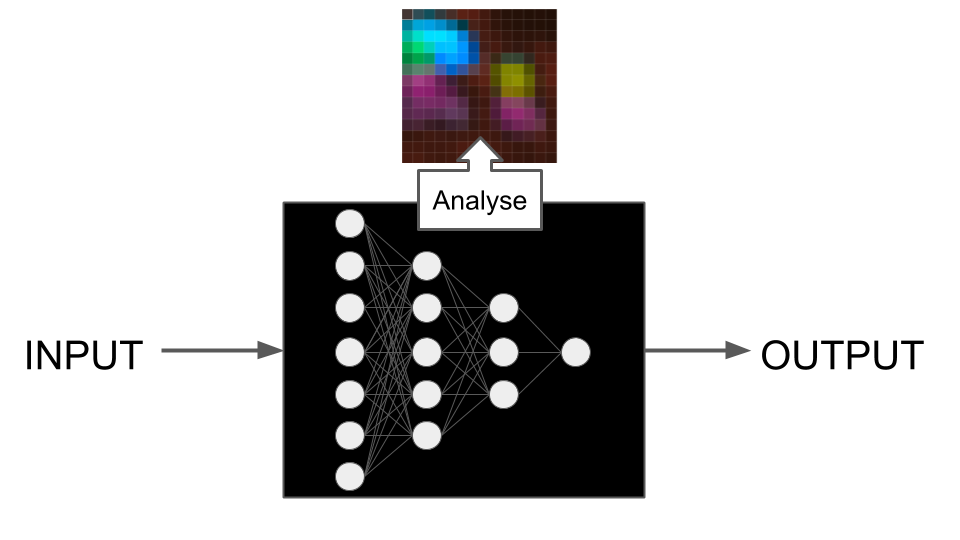}
    \end{minipage}\hfill
    \begin{minipage}{0.45\textwidth}
        \centering
        \includegraphics[width=0.9\textwidth]{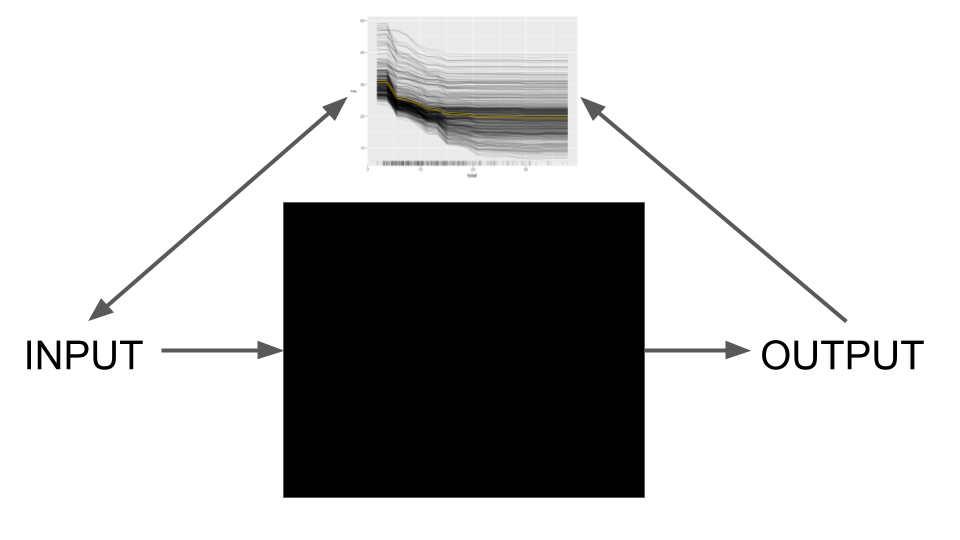}
    \end{minipage}
    \label{fig:iml-type}
    \caption{Some IML approaches work by assigning meaning to individual model components (left), some by analyzing the model predictions for perturbations of the data (right). The surrogate approach, a mixture of the two other approaches, approximates the ML model using (perturbed) data and then analyzes the components of the interpretable surrogate model.}
\end{figure}

We distinguish IML methods by whether they analyze model components, model sensitivity\footnote{Not to be confused with the research field of sensitivity analysis, which studies the uncertainty of outputs in mathematical models and systems. There are methodological overlaps (e.g., Shapley values), but also differences in methods and how input data distributions are handled.}, or surrogate models, illustrated in Figure~\ref{fig:iml-type}.\footnote{Some surveys distinguish between \textit{ante-hoc (or transparent design, white-box models, inherently interpretable model)} and \textit{post-hoc} IML method, depending on whether interpretability is considered at model design and training or after training, leaving the (black-box) model unchanged. Another category separates model-agnostic and model-specific methods.}

\subsection{Analyzing Components of Interpretable Models}

In order to analyze components of a model, it needs to be decomposable into parts that we can interpret individually. 
However, it is not necessarily required that the user understands the model in its entirety (simulatability \cite{murdoch2019definitions}).
Component analysis is always model-specific, because it is tied to the structure of the model.

Inherently interpretable models are models with (learned) structures and (learned) parameters which can be assigned a certain interpretation.
In this context, linear regression models, decision trees and decision rules are considered to be interpretable \cite{freitas2014comprehensible,huysmans2011empirical}.
Linear regression models can be interpreted by analyzing components:
The model structure, a weighted sum of features, allows to interpret the weights as the effects that the features have on the prediction.

Decision trees and other rule-based ML models have a learned structure (e.g.,\enquote{IF feature $x_1 > 0$ and feature $x_2 \in \{A,B\}$, THEN predict 0.6}).
We can interpret the learned structure to trace how the model makes predictions.

This only works up to a certain point in high-dimensional scenarios.
Linear regression models with hundreds of features and complex interaction terms or deep decision trees are not that interpretable anymore.
Some approaches aim to reduce the parts to be interpreted.
For example, LASSO \cite{santosa1986linear,tibshirani1996regression} shrinks the coefficients in a linear model so that many of them become zero, and pruning techniques shorten trees.

\subsection{Analyzing Components of More Complex Models}

With a bit more effort, we can also analyze components of more complex black-box models.
\footnote{This blurs the line between an \enquote{inherently interpretable} and a \enquote{black-box} model.}
For example, the abstract features learned by a deep convolutional neural network (CNN) can be visualized by finding or generating images that activate a feature map of the CNN \cite{olah2017feature}.
For the random forest, the minimal depth distribution \cite{randomForestExplainer,ishwaran2010high} and the Gini importance \cite{breiman2001random} analyze the structure of the trees of the forest and can be used to quantify feature importance.
Some approaches aim to make the parts of a model more interpretable with, for example, a monotonicity constraint
\cite{sill1998monotonic} or a modified loss function for disentangling concepts learned by a convolutional neural network \cite{zhang2018interpretable}.

If an ML algorithm is well understood and frequently used in a community, like random forests in ecology research \cite{cutler2007random}, model component analysis can be the correct tool, but it has the obvious disadvantage that it is tied to that specific model. 
And it does not combine well with the common model selection approach in ML, where one usually searches over a large class of different ML models via cross-validation.

\subsection{Explaining Individual Predictions}

Methods that study the sensitivity of an ML model are mostly model-agnostic and work by manipulating input data and analyzing the respective model predictions.
These IML methods often treat the ML model as a closed system that receives feature values as an input and produces a prediction as output.
We distinguish between local and global explanations.

Local methods explain individual predictions of ML models.
Local explanation methods have received much attention and there has been a lot of innovation in the last years.
Popular local IML methods are Shapley values \cite{lundberg2017unified,vstrumbelj2014explaining} and counterfactual explanations \cite{wachter2017counterfactual,dandl2020multi,mothilal2020explaining,tolomei2017interpretable,ustun2019actionable}.
Counterfactual explanations explain predictions in the form of what-if scenarios, which builds on a rich tradition in philosophy \cite{starr2019counterfactuals}.
According to findings in the social sciences \cite{miller2019explanation}, counterfactual explanations are \enquote{good} explanations because they are contrastive and focus on a few reasons.
A different approach originates from collaborative game theory:
The Shapley values \cite{shapley1953value} provide an answer on how to fairly share a payout among the players of a collaborative game.
The collaborative game idea can be applied to ML where features (i.e., the players) collaborate to make a prediction (i.e., the payout) \cite{vstrumbelj2014explaining,lundberg2017unified,lundberg2018consistent}.

Some IML methods rely on model-specific knowledge to analyze how changes in the input features change the output.
Saliency maps, an interpretation method specific for CNNs, make use of the network gradients to explain individual classifications.
The explanations are in the form of heatmaps that show how changing a pixel can change the classification.
The saliency map methods differ in how they backpropagate \cite{sundararajan2017axiomatic,lundberg2017unified,montavon2017explaining,simonyan2013deep,shrikumar2016not}.
Additionally, model-agnostic versions \cite{ribeiro2016should,lundberg2017unified,zeiler2014visualizing} exist for analyzing image classifiers.

\subsection{Explaining Global Model Behavior}

Global model-agnostic explanation methods are used to explain the expected model behavior, i.e., how the model behaves on average for a given dataset.
A useful distinction of global explanations are feature importance and feature effect.

Feature importance ranks features based on how relevant they were for the prediction.
Permutation feature importance \cite{fisher2019all,casalicchio2018visualizing} is a popular importance measure, originally suggested for random forests \cite{breiman2001random}.
Some importance measures rely on removing features from the training data and retraining the model \cite{lei2018distribution}.
An alternative are variance-based measures \cite{greenwell2018simple}.
See \cite{wei2015variable} for an overview of importance measures.

The feature effect expresses how a change in a feature changes the predicted outcome.
Popular feature effect plots are partial dependence plots \cite{friedman2001greedy}, individual conditional expectation curves \cite{goldstein2015peeking}, accumulated local effect plots \cite{apley2016visualizing}, and the functional ANOVA \cite{hooker2007generalized}.
Analyzing influential data instances, inspired by statistics, provides a different view into the model and describes how influential a data point was for a prediction \cite{koh2017understanding}.

\subsection{Surrogate Models}

Surrogate models\footnote{Surrogate models are related to knowledge distillation and the teacher-student model.} are interpretable models designed to \enquote{copy} the behavior of the ML model.
The surrogate approach treats the ML model as a black-box and only requires the input and output data of the ML model (similar to sensitivity analysis) to train a surrogate ML model.
However, the interpretation is based on analyzing components of the interpretable surrogate model.
Many IML methods are surrogate model approaches \cite{puri2017magix,molnar2019,ming2018rulematrix,ribeiro2016should,frosst2017distilling,bastani2017interpreting,craven1996extracting,krishnan2017palm} and differ, e.g., in the targeted ML model, the data sampling strategy, or the interpretable model that is used.
There are also methods for extracting, e.g., decision rules from specific models based on their internal components such as neural network weights \cite{andrews1995survey,augasta2012rule}.
LIME \cite{ribeiro2016should} is an example of a local surrogate method that explains individual predictions by learning an interpretable model with data in proximity to the data point to be explained.
Numerous extensions of LIME exist, which try to fix issues with the original method, extend it to other tasks and data, or analyze its properties \cite{hu2018locally,rabold2018explaining,rabold2019enriching,visani2020optilime,haunschmid2020audiolime,rahnama2019study,shankaranarayana2019alime,botari2020melime}.

\section{Challenges}

This section presents an incomplete overview of challenges for IML, mostly based on \cite{molnar2020pitfalls}.

\subsection{Statistical Uncertainty and Inference}

Many IML methods such as permutation feature importance or Shapley values provide explanations without quantifying the uncertainty of the explanation.
The model itself, but also its explanations, are computed from data and hence are subject to uncertainty.
First research is working towards quantifying uncertainty of explanations, for example, for feature importance \cite{watson2019testing,fisher2019all,altmann2010permutation}, layer-wise relevance propagation \cite{fabi2020feature}, and Shapley values \cite{williamson2020efficient}.

In order to infer meaningful properties of the underlying data generating process, we have to make structural or distributional assumptions.
Whether it is a classical statistical model, an ML algorithm or an IML procedure, these assumptions should be clearly stated and we need better diagnostic tools to test them.
If we want to prevent statistical testing problems such as p-hacking \cite{head2015extent} to reappear in IML, we have to become more rigorous in studying and quantifying the uncertainty of IML methods.
For example, most IML methods for feature importance are not adapted for multiple testing, which is a classic mistake in a statistical analysis.

\subsection{Causal Interpretation}
Ideally, a model should reflect the true causal structure of its underlying phenomena, to enable
causal interpretations.
Arguably, causal interpretation is usually the goal of modeling if ML is used in science.
But most statistical learning procedures reflect mere correlation structures between features and analyze the surface of the data generation process instead of its true inherent structure. 
Such causal structures would also make models more robust against adversarial attacks \cite{scholkopf2019causality,freiesleben2020counterfactual}, and more useful when used as a basis for decision making.
Unfortunately, predictive performance and causality can be conflicting goals.
For example, today's weather directly causes tomorrow's weather, but we might only have access to the feature \enquote{wet ground}.
Using \enquote{wet ground} in the prediction model for \enquote{tomorrow's weather} is useful as it has information about \enquote{today's weather}, but we are not allowed to interpret it causally, because the confounder \enquote{today's weather} is missing from the ML model.
Further research is needed to understand when we are allowed to make causal interpretations of an ML model.
First steps have been made for permutation feature importance \cite{konig2020relative} and Shapley values \cite{ma2020predictive}.

\subsection{Feature Dependence}

Feature dependence introduces problems with attribution and extrapolation.
Attribution of importance and effects of features becomes difficult when features are, for example, correlated and therefore share information.
Correlated features in random forests are preferred and attributed a higher importance \cite{strobl2008conditional,hooker2019please}.
Many sensitivity analysis based methods permute features.
When the permuted feature has some dependence with another feature, this association is broken and the resulting data points extrapolate to areas outside the distribution.
The ML model was never trained on such combinations and will likely not be confronted with similar data points in an application.
Therefore, extrapolation can cause misleading interpretations.
There have been attempts to \enquote{fix} permutation-based methods, by using a conditional permutation scheme that respects the joint distribution of the data  \cite{molnar2020model,strobl2008conditional,fisher2019all,hooker2019please}.
The change from unconditional to conditional permutation changes the respective interpretation method \cite{molnar2020model,apley2016visualizing}, or, in worst case, can break it \cite{janzing2019feature,sundararajan2019many,kumar2020problems}.

\subsection{Definition of Interpretability}
A lack of definition for the term "interpretability" is a common critique of the field \cite{lipton2018mythos,doshi2017towards}.
How can we decide if a new method explains ML models better without a satisfying definition of interpretability?
To evaluate the predictive performance of an ML model, we simply compute the prediction error on test data given the groundtruth label.
To evaluate the interpretability of that same ML model is more difficult.
We do not know what the groundtruth explanation looks like and have no straightforward way to quantify how interpretable a model is or how correct an explanation is.
Instead of having one groundtruth explanation, various quantifiable aspects of interpretability are emerging \cite{poursabzi2018manipulating,philipp2018measuring,molnar2019quantifying,hauenstein2018computing,zhou2018measuring,akaike1998information,schwarz1978estimating,poursabzi2018manipulating,dhurandhar2017tip,friedler2019assessing}.

The two main ways of evaluating interpretability are objective evaluations, which are mathematically quantifiable metrics, and {human-centered evaluations}, which involve studies with either domain experts or lay persons.
Examples of aspects of interpretability are sparsity, interaction strength, fidelity (how well an explanation approximates the ML model), sensitivity to perturbations, and a user's ability to run a model on a given input (simulatability).
The challenge ahead remains to establish a best practice on how to evaluate interpretation methods and the explanations they produce.
Here, we should also look at the field of human-computer interaction.

\subsection{More Challenges Ahead}
We focused mainly on the methodological, mathematical challenges in a rather static setting, where a trained ML model and the data are assumed as given and fixed.
But ML models are usually not used in a static and isolated way, but are embedded in some process or product, and interact with people.
A more dynamic and holistic view of the entire process, from data collection to the final consumption of the explained prediction is needed.
This includes thinking how to explain predictions to individuals with diverse knowledge and backgrounds and about the need of interpretability on the level of an institution or society in general.
This covers a wide range of fields, such as human-computer interaction, psychology and sociology.
To solve the challenges ahead, we believe that the field has to reach out horizontally -- to other domains -- and vertically -- drawing from the rich research in statistics and computer science.

\vskip 0.2in
\bibliographystyle{splncs04}
\bibliography{Bib}

\begin{thebibliography}{100}
\providecommand{\url}[1]{\texttt{#1}}
\providecommand{\urlprefix}{URL }
\providecommand{\doi}[1]{https://doi.org/#1}

\bibitem{adadi2018peeking}
Adadi, A., Berrada, M.: Peeking inside the black-box: A survey on explainable
  artificial intelligence (xai). IEEE Access  \textbf{6},  52138--52160 (2018)

\bibitem{adebayo2018sanity}
Adebayo, J., Gilmer, J., Muelly, M., Goodfellow, I., Hardt, M., Kim, B.: Sanity
  checks for saliency maps. In: Advances in Neural Information Processing
  Systems. pp. 9505--9515 (2018)

\bibitem{akaike1998information}
Akaike, H.: Information theory and an extension of the maximum likelihood
  principle. In: Selected papers of {H}irotugu {A}kaike, pp. 199--213. Springer
  (1998)

\bibitem{altmann2010permutation}
Altmann, A., Tolo{\c{s}}i, L., Sander, O., Lengauer, T.: Permutation
  importance: a corrected feature importance measure. Bioinformatics
  \textbf{26}(10),  1340--1347 (2010)

\bibitem{andrews1995survey}
Andrews, R., Diederich, J., Tickle, A.B.: Survey and critique of techniques for
  extracting rules from trained artificial neural networks. Knowledge-based
  systems  \textbf{8}(6),  373--389 (1995)

\bibitem{anjomshoae2019explainable}
Anjomshoae, S., Najjar, A., Calvaresi, D., Fr{\"a}mling, K.: Explainable agents
  and robots: Results from a systematic literature review. In: 18th
  International Conference on Autonomous Agents and Multiagent Systems (AAMAS
  2019), Montreal, Canada, May 13--17, 2019. pp. 1078--1088. International
  Foundation for Autonomous Agents and Multiagent Systems (2019)

\bibitem{apley2016visualizing}
Apley, D.W., Zhu, J.: Visualizing the effects of predictor variables in black
  box supervised learning models. arXiv preprint arXiv:1612.08468  (2016)

\bibitem{arya2020ai}
Arya, V., Bellamy, R.K., Chen, P.Y., Dhurandhar, A., Hind, M., Hoffman, S.C.,
  Houde, S., Liao, Q.V., Luss, R., Mojsilovic, A., et~al.: {AI} explainability
  360: An extensible toolkit for understanding data and machine learning
  models. Journal of Machine Learning Research  \textbf{21}(130), ~1--6 (2020)

\bibitem{augasta2012rule}
Augasta, M.G., Kathirvalavakumar, T.: Rule extraction from neural networks—a
  comparative study. In: International Conference on Pattern Recognition,
  Informatics and Medical Engineering (PRIME-2012). pp. 404--408. IEEE (2012)

\bibitem{bastani2017interpreting}
Bastani, O., Kim, C., Bastani, H.: Interpreting blackbox models via model
  extraction. arXiv preprint arXiv:1705.08504  (2017)

\bibitem{biecek2018dalex}
Biecek, P.: {DALEX}: explainers for complex predictive models in r. The Journal
  of Machine Learning Research  \textbf{19}(1),  3245--3249 (2018)

\bibitem{botari2020melime}
Botari, T., Hvilsh{\o}j, F., Izbicki, R., de~Carvalho, A.C.: {MeLIME}:
  Meaningful local explanation for machine learning models. arXiv preprint
  arXiv:2009.05818  (2020)

\bibitem{breiman2001random}
Breiman, L.: Random forests. Machine learning  \textbf{45}(1),  5--32 (2001)

\bibitem{caruana2015intelligible}
Caruana, R., Lou, Y., Gehrke, J., Koch, P., Sturm, M., Elhadad, N.:
  Intelligible models for healthcare: Predicting pneumonia risk and hospital
  30-day readmission. In: Proceedings of the 21th ACM SIGKDD international
  conference on knowledge discovery and data mining. pp. 1721--1730 (2015)

\bibitem{carvalho2019machine}
Carvalho, D.V., Pereira, E.M., Cardoso, J.S.: Machine learning
  interpretability: A survey on methods and metrics. Electronics
  \textbf{8}(8), ~832 (2019)

\bibitem{casalicchio2018visualizing}
Casalicchio, G., Molnar, C., Bischl, B.: Visualizing the feature importance for
  black box models. In: Joint European Conference on Machine Learning and
  Knowledge Discovery in Databases. pp. 655--670. Springer (2018)

\bibitem{chromik2020taxonomy}
Chromik, M., Schuessler, M.: A taxonomy for human subject evaluation of
  black-box explanations in {XAI}. In: ExSS-ATEC@ IUI (2020)

\bibitem{craven1996extracting}
Craven, M., Shavlik, J.W.: Extracting tree-structured representations of
  trained networks. In: Advances in neural information processing systems. pp.
  24--30 (1996)

\bibitem{cutler2007random}
Cutler, D.R., Edwards~Jr, T.C., Beard, K.H., Cutler, A., Hess, K.T., Gibson,
  J., Lawler, J.J.: Random forests for classification in ecology. Ecology
  \textbf{88}(11),  2783--2792 (2007)

\bibitem{dandl2020multi}
Dandl, S., Molnar, C., Binder, M., Bischl, B.: Multi-objective counterfactual
  explanations. arXiv preprint arXiv:2004.11165  (2020)

\bibitem{dhurandhar2017tip}
Dhurandhar, A., Iyengar, V., Luss, R., Shanmugam, K.: {TIP:} typifying the
  interpretability of procedures. arXiv preprint arXiv:1706.02952  (2017)

\bibitem{doshi2017towards}
Doshi-Velez, F., Kim, B.: Towards a rigorous science of interpretable machine
  learning. arXiv preprint arXiv:1702.08608  (2017)

\bibitem{du2019techniques}
Du, M., Liu, N., Hu, X.: Techniques for interpretable machine learning.
  Communications of the ACM  \textbf{63}(1),  68--77 (2019)

\bibitem{fabi2020feature}
Fabi, K., Schneider, J.: On feature relevance uncertainty: A {M}onte {C}arlo
  dropout sampling approach. arXiv preprint arXiv:2008.01468  (2020)

\bibitem{fahrmeir2013multivariate}
Fahrmeir, L., Tutz, G.: Multivariate statistical modelling based on generalized
  linear models. Springer Science \& Business Media (2013)

\bibitem{fasiolo2020scalable}
Fasiolo, M., Nedellec, R., Goude, Y., Wood, S.N.: Scalable visualization
  methods for modern generalized additive models. Journal of computational and
  Graphical Statistics  \textbf{29}(1),  78--86 (2020)

\bibitem{fasiolo2020fast}
Fasiolo, M., Wood, S.N., Zaffran, M., Nedellec, R., Goude, Y.: Fast calibrated
  additive quantile regression. Journal of the American Statistical Association
  pp. 1--11 (2020)

\bibitem{fisher2019all}
Fisher, A., Rudin, C., Dominici, F.: All models are wrong, but many are useful:
  Learning a variable’s importance by studying an entire class of prediction
  models simultaneously. Journal of Machine Learning Research
  \textbf{20}(177),  1--81 (2019)

\bibitem{freiesleben2020counterfactual}
Freiesleben, T.: Counterfactual explanations \& adversarial examples--common
  grounds, essential differences, and potential transfers. arXiv preprint
  arXiv:2009.05487  (2020)

\bibitem{freitas2014comprehensible}
Freitas, A.A.: Comprehensible classification models: a position paper. ACM
  SIGKDD explorations newsletter  \textbf{15}(1),  1--10 (2014)

\bibitem{friedler2019assessing}
Friedler, S.A., Roy, C.D., Scheidegger, C., Slack, D.: Assessing the local
  interpretability of machine learning models. arXiv preprint arXiv:1902.03501
  (2019)

\bibitem{friedman2001greedy}
Friedman, J.H.: Greedy function approximation: a gradient boosting machine.
  Annals of statistics pp. 1189--1232 (2001)

\bibitem{friedman2008predictive}
Friedman, J.H., Popescu, B.E., et~al.: Predictive learning via rule ensembles.
  The Annals of Applied Statistics  \textbf{2}(3),  916--954 (2008)

\bibitem{frosst2017distilling}
Frosst, N., Hinton, G.: Distilling a neural network into a soft decision tree.
  arXiv preprint arXiv:1711.09784  (2017)

\bibitem{furnkranz2012foundations}
F{\"u}rnkranz, J., Gamberger, D., Lavra{\v{c}}, N.: Foundations of rule
  learning. Springer Science \& Business Media (2012)

\bibitem{gade2019explainable}
Gade, K., Geyik, S.C., Kenthapadi, K., Mithal, V., Taly, A.: Explainable {AI}
  in industry. In: Proceedings of the 25th ACM SIGKDD International Conference
  on Knowledge Discovery \& Data Mining. pp. 3203--3204 (2019)

\bibitem{gauss1809theoria}
Gauss, C.F.: Theoria motus corporum coelestium in sectionibus conicis solem
  ambientium, vol.~7. Perthes et Besser (1809)

\bibitem{gelman2006data}
Gelman, A., Hill, J.: Data analysis using regression and
  multilevel/hierarchical models. Cambridge university press (2006)

\bibitem{goldstein2015peeking}
Goldstein, A., Kapelner, A., Bleich, J., Pitkin, E.: Peeking inside the black
  box: Visualizing statistical learning with plots of individual conditional
  expectation. Journal of Computational and Graphical Statistics
  \textbf{24}(1),  44--65 (2015)

\bibitem{greenwell2018simple}
Greenwell, B.M., Boehmke, B.C., McCarthy, A.J.: A simple and effective
  model-based variable importance measure. arXiv preprint arXiv:1805.04755
  (2018)

\bibitem{guidotti2018survey}
Guidotti, R., Monreale, A., Ruggieri, S., Turini, F., Giannotti, F., Pedreschi,
  D.: A survey of methods for explaining black box models. ACM computing
  surveys (CSUR)  \textbf{51}(5),  1--42 (2018)

\bibitem{hall2019systematic}
Hall, M., Harborne, D., Tomsett, R., Galetic, V., Quintana-Amate, S., Nottle,
  A., Preece, A.: A systematic method to understand requirements for
  explainable {AI}({XAI}) systems. In: Proceedings of the IJCAI Workshop on
  eXplainable Artificial Intelligence (XAI 2019), Macau, China (2019)

\bibitem{hall2017machine}
Hall, P., Gill, N., Kurka, M., Phan, W.: Machine learning interpretability with
  h2o driverless {AI}. H2O. ai. URL: http://docs. h2o.
  ai/driverless-ai/latest-stable/docs/booklets/MLIBooklet. pdf  (2017)

\bibitem{hapfelmeier2014new}
Hapfelmeier, A., Hothorn, T., Ulm, K., Strobl, C.: A new variable importance
  measure for random forests with missing data. Statistics and Computing
  \textbf{24}(1),  21--34 (2014)

\bibitem{hastie1990generalized}
Hastie, T.J., Tibshirani, R.J.: Generalized additive models, vol.~43. CRC press
  (1990)

\bibitem{hauenstein2018computing}
Hauenstein, S., Wood, S.N., Dormann, C.F.: Computing {AIC} for black-box models
  using generalized degrees of freedom: A comparison with cross-validation.
  Communications in Statistics-Simulation and Computation  \textbf{47}(5),
  1382--1396 (2018)

\bibitem{haunschmid2020audiolime}
Haunschmid, V., Manilow, E., Widmer, G.: {audioLIME}: Listenable explanations
  using source separation. arXiv preprint arXiv:2008.00582  (2020)

\bibitem{head2015extent}
Head, M.L., Holman, L., Lanfear, R., Kahn, A.T., Jennions, M.D.: The extent and
  consequences of p-hacking in science. PLoS Biol  \textbf{13}(3),  e1002106
  (2015)

\bibitem{hoffman2018metrics}
Hoffman, R.R., Mueller, S.T., Klein, G., Litman, J.: Metrics for explainable
  {AI}: Challenges and prospects. arXiv preprint arXiv:1812.04608  (2018)

\bibitem{hooker2007generalized}
Hooker, G.: Generalized functional anova diagnostics for high-dimensional
  functions of dependent variables. Journal of Computational and Graphical
  Statistics  \textbf{16}(3),  709--732 (2007)

\bibitem{hooker2019please}
Hooker, G., Mentch, L.: Please stop permuting features: An explanation and
  alternatives. arXiv preprint arXiv:1905.03151  (2019)

\bibitem{hothorn2015ctree}
Hothorn, T., Hornik, K., Zeileis, A.: ctree: Conditional inference trees. The
  Comprehensive R Archive Network  \textbf{8} (2015)

\bibitem{hu2018locally}
Hu, L., Chen, J., Nair, V.N., Sudjianto, A.: Locally interpretable models and
  effects based on supervised partitioning ({LIME-SUP}). arXiv preprint
  arXiv:1806.00663  (2018)

\bibitem{huysmans2011empirical}
Huysmans, J., Dejaeger, K., Mues, C., Vanthienen, J., Baesens, B.: An empirical
  evaluation of the comprehensibility of decision table, tree and rule based
  predictive models. Decision Support Systems  \textbf{51}(1),  141--154 (2011)

\bibitem{ishwaran2010high}
Ishwaran, H., Kogalur, U.B., Gorodeski, E.Z., Minn, A.J., Lauer, M.S.:
  High-dimensional variable selection for survival data. Journal of the
  American Statistical Association  \textbf{105}(489),  205--217 (2010)

\bibitem{ishwaran2007variable}
Ishwaran, H., et~al.: Variable importance in binary regression trees and
  forests. Electronic Journal of Statistics  \textbf{1},  519--537 (2007)

\bibitem{janzing2019feature}
Janzing, D., Minorics, L., Bl{\"o}baum, P.: Feature relevance quantification in
  explainable {AI}: A causality problem. arXiv preprint arXiv:1910.13413
  (2019)

\bibitem{klaise2020alibi}
Klaise, J., Van~Looveren, A., Vacanti, G., Coca, A.: Alibi: Algorithms for
  monitoring and explaining machine learning models. URL https://github.
  com/SeldonIO/alibi  (2020)

\bibitem{koh2017understanding}
Koh, P.W., Liang, P.: Understanding black-box predictions via influence
  functions. arXiv preprint arXiv:1703.04730  (2017)

\bibitem{konig2020relative}
K{\"o}nig, G., Molnar, C., Bischl, B., Grosse-Wentrup, M.: Relative feature
  importance. arXiv preprint arXiv:2007.08283  (2020)

\bibitem{krishnan2017palm}
Krishnan, S., Wu, E.: Palm: Machine learning explanations for iterative
  debugging. In: Proceedings of the 2nd Workshop on Human-In-the-Loop Data
  Analytics. pp.~1--6 (2017)

\bibitem{kumar2020problems}
Kumar, I.E., Venkatasubramanian, S., Scheidegger, C., Friedler, S.: Problems
  with {S}hapley-value-based explanations as feature importance measures. arXiv
  preprint arXiv:2002.11097  (2020)

\bibitem{laugel2019dangers}
Laugel, T., Lesot, M.J., Marsala, C., Renard, X., Detyniecki, M.: The dangers
  of post-hoc interpretability: Unjustified counterfactual explanations. arXiv
  preprint arXiv:1907.09294  (2019)

\bibitem{legendre1805nouvelles}
Legendre, A.M.: Nouvelles m{\'e}thodes pour la d{\'e}termination des orbites
  des com{\`e}tes. F. Didot (1805)

\bibitem{lei2018distribution}
Lei, J., G’Sell, M., Rinaldo, A., Tibshirani, R.J., Wasserman, L.:
  Distribution-free predictive inference for regression. Journal of the
  American Statistical Association  \textbf{113}(523),  1094--1111 (2018)

\bibitem{letham2015interpretable}
Letham, B., Rudin, C., McCormick, T.H., Madigan, D., et~al.: Interpretable
  classifiers using rules and bayesian analysis: Building a better stroke
  prediction model. The Annals of Applied Statistics  \textbf{9}(3),
  1350--1371 (2015)

\bibitem{lipton2018mythos}
Lipton, Z.C.: The mythos of model interpretability. Queue  \textbf{16}(3),
  31--57 (2018)

\bibitem{lundberg2018consistent}
Lundberg, S.M., Erion, G.G., Lee, S.I.: Consistent individualized feature
  attribution for tree ensembles. arXiv preprint arXiv:1802.03888  (2018)

\bibitem{lundberg2017unified}
Lundberg, S.M., Lee, S.I.: A unified approach to interpreting model
  predictions. In: Advances in neural information processing systems. pp.
  4765--4774 (2017)

\bibitem{ma2020predictive}
Ma, S., Tourani, R.: Predictive and causal implications of using {S}hapley
  value for model interpretation. In: Proceedings of the 2020 KDD Workshop on
  Causal Discovery. pp. 23--38. PMLR (2020)

\bibitem{miller2019explanation}
Miller, T.: Explanation in artificial intelligence: Insights from the social
  sciences. Artificial Intelligence  \textbf{267},  1--38 (2019)

\bibitem{ming2018rulematrix}
Ming, Y., Qu, H., Bertini, E.: Rulematrix: Visualizing and understanding
  classifiers with rules. IEEE transactions on visualization and computer
  graphics  \textbf{25}(1),  342--352 (2018)

\bibitem{mohseni2018human}
Mohseni, S., Ragan, E.D.: A human-grounded evaluation benchmark for local
  explanations of machine learning. arXiv preprint arXiv:1801.05075  (2018)

\bibitem{mohseni2018multidisciplinary}
Mohseni, S., Zarei, N., Ragan, E.D.: A multidisciplinary survey and framework
  for design and evaluation of explainable {AI} systems. arXiv pp. arXiv--1811
  (2018)

\bibitem{molnar2019}
Molnar, C.: Interpretable Machine Learning (2019),
  \url{https://christophm.github.io/interpretable-ml-book/}

\bibitem{iml}
Molnar, C., Bischl, B., Casalicchio, G.: iml: An {R} package for interpretable
  machine learning. JOSS  \textbf{3}(26), ~786 (2018)

\bibitem{molnar2019quantifying}
Molnar, C., Casalicchio, G., Bischl, B.: Quantifying model complexity via
  functional decomposition for better post-hoc interpretability. In: Joint
  European Conference on Machine Learning and Knowledge Discovery in Databases.
  pp. 193--204. Springer (2019)

\bibitem{molnar2020model}
Molnar, C., K{\"o}nig, G., Bischl, B., Casalicchio, G.: Model-agnostic feature
  importance and effects with dependent features--a conditional subgroup
  approach. arXiv preprint arXiv:2006.04628  (2020)

\bibitem{molnar2020pitfalls}
Molnar, C., K{\"o}nig, G., Herbinger, J., Freiesleben, T., Dandl, S.,
  Scholbeck, C.A., Casalicchio, G., Grosse-Wentrup, M., Bischl, B.: Pitfalls to
  avoid when interpreting machine learning models. arXiv preprint
  arXiv:2007.04131  (2020)

\bibitem{montavon2017explaining}
Montavon, G., Lapuschkin, S., Binder, A., Samek, W., M{\"u}ller, K.R.:
  Explaining nonlinear classification decisions with deep taylor decomposition.
  Pattern Recognition  \textbf{65},  211--222 (2017)

\bibitem{mothilal2020explaining}
Mothilal, R.K., Sharma, A., Tan, C.: Explaining machine learning classifiers
  through diverse counterfactual explanations. In: Proceedings of the 2020
  Conference on Fairness, Accountability, and Transparency. pp. 607--617 (2020)

\bibitem{murdoch2019definitions}
Murdoch, W.J., Singh, C., Kumbier, K., Abbasi-Asl, R., Yu, B.: Definitions,
  methods, and applications in interpretable machine learning. Proceedings of
  the National Academy of Sciences  \textbf{116}(44),  22071--22080 (2019)

\bibitem{nori2019interpretml}
Nori, H., Jenkins, S., Koch, P., Caruana, R.: Interpretml: A unified framework
  for machine learning interpretability. arXiv preprint arXiv:1909.09223
  (2019)

\bibitem{olah2017feature}
Olah, C., Mordvintsev, A., Schubert, L.: Feature visualization. Distill
  (2017). \doi{10.23915/distill.00007},
  https://distill.pub/2017/feature-visualization

\bibitem{randomForestExplainer}
Paluszynska, A., Biecek, P., Jiang, Y.: randomForestExplainer: Explaining and
  Visualizing Random Forests in Terms of Variable Importance (2020),
  \url{https://CRAN.R-project.org/package=randomForestExplainer}, r package
  version 0.10.1

\bibitem{philipp2018measuring}
Philipp, M., Rusch, T., Hornik, K., Strobl, C.: Measuring the stability of
  results from supervised statistical learning. Journal of Computational and
  Graphical Statistics  \textbf{27}(4),  685--700 (2018)

\bibitem{poursabzi2018manipulating}
Poursabzi-Sangdeh, F., Goldstein, D.G., Hofman, J.M., Vaughan, J.W., Wallach,
  H.: Manipulating and measuring model interpretability. arXiv preprint
  arXiv:1802.07810  (2018)

\bibitem{preece2018stakeholders}
Preece, A., Harborne, D., Braines, D., Tomsett, R., Chakraborty, S.:
  Stakeholders in explainable {AI}. arXiv preprint arXiv:1810.00184  (2018)

\bibitem{puri2017magix}
Puri, N., Gupta, P., Agarwal, P., Verma, S., Krishnamurthy, B.: Magix: Model
  agnostic globally interpretable explanations. arXiv preprint arXiv:1706.07160
   (2017)

\bibitem{quetelet1827recherches}
Quetelet, L.A.J.: Recherches sur la population, les naissances, les
  d{\'e}c{\`e}s, les prisons, les d{\'e}p{\^o}ts de mendicit{\'e}, etc. dans le
  royaume des Pays-Bas (1827)

\bibitem{Rlang}
{R Core Team}: {R}: A Language and Environment for Statistical Computing. R
  Foundation for Statistical Computing, Vienna, Austria (2020),
  \url{https://www.R-project.org/}

\bibitem{rabold2019enriching}
Rabold, J., Deininger, H., Siebers, M., Schmid, U.: Enriching visual with
  verbal explanations for relational concepts--combining {LIME} with {Aleph}.
  In: Joint European Conference on Machine Learning and Knowledge Discovery in
  Databases. pp. 180--192. Springer (2019)

\bibitem{rabold2018explaining}
Rabold, J., Siebers, M., Schmid, U.: Explaining black-box classifiers with
  ilp--empowering {LIME} with aleph to approximate non-linear decisions with
  relational rules. In: International Conference on Inductive Logic
  Programming. pp. 105--117. Springer (2018)

\bibitem{rahnama2019study}
Rahnama, A.H.A., Bostr{\"o}m, H.: A study of data and label shift in the {LIME}
  framework. arXiv preprint arXiv:1910.14421  (2019)

\bibitem{ribeiro2016should}
Ribeiro, M.T., Singh, S., Guestrin, C.: " why should i trust you?" explaining
  the predictions of any classifier. In: Proceedings of the 22nd ACM SIGKDD
  international conference on knowledge discovery and data mining. pp.
  1135--1144 (2016)

\bibitem{rosenfeld2019explainability}
Rosenfeld, A., Richardson, A.: Explainability in human--agent systems.
  Autonomous Agents and Multi-Agent Systems  \textbf{33}(6),  673--705 (2019)

\bibitem{samek2019towards}
Samek, W., M{\"u}ller, K.R.: Towards explainable artificial intelligence. In:
  Explainable AI: interpreting, explaining and visualizing deep learning, pp.
  5--22. Springer (2019)

\bibitem{santosa1986linear}
Santosa, F., Symes, W.W.: Linear inversion of band-limited reflection
  seismograms. SIAM Journal on Scientific and Statistical Computing
  \textbf{7}(4),  1307--1330 (1986)

\bibitem{schapire1990strength}
Schapire, R.E.: The strength of weak learnability. Machine learning
  \textbf{5}(2),  197--227 (1990)

\bibitem{schmidhuber2015deep}
Schmidhuber, J.: Deep learning in neural networks: An overview. Neural networks
   \textbf{61},  85--117 (2015)

\bibitem{scholkopf2019causality}
Sch{\"o}lkopf, B.: Causality for machine learning. arXiv preprint
  arXiv:1911.10500  (2019)

\bibitem{schwarz1978estimating}
Schwarz, G., et~al.: Estimating the dimension of a model. The annals of
  statistics  \textbf{6}(2),  461--464 (1978)

\bibitem{shankaranarayana2019alime}
Shankaranarayana, S.M., Runje, D.: {ALIME}: Autoencoder based approach for
  local interpretability. In: International Conference on Intelligent Data
  Engineering and Automated Learning. pp. 454--463. Springer (2019)

\bibitem{shapley1953value}
Shapley, L.S.: A value for n-person games. Contributions to the Theory of Games
   \textbf{2}(28),  307--317 (1953)

\bibitem{shrikumar2016not}
Shrikumar, A., Greenside, P., Shcherbina, A., Kundaje, A.: Not just a black
  box: Learning important features through propagating activation differences.
  arXiv preprint arXiv:1605.01713  (2016)

\bibitem{sill1998monotonic}
Sill, J.: Monotonic networks. In: Advances in neural information processing
  systems. pp. 661--667 (1998)

\bibitem{simonyan2013deep}
Simonyan, K., Vedaldi, A., Zisserman, A.: Deep inside convolutional networks:
  Visualising image classification models and saliency maps. arXiv preprint
  arXiv:1312.6034  (2013)

\bibitem{starr2019counterfactuals}
Starr, W.: Counterfactuals  (2019)

\bibitem{stigler1986history}
Stigler, S.M.: The history of statistics: The measurement of uncertainty before
  1900. Harvard University Press (1986)

\bibitem{strobl2008conditional}
Strobl, C., Boulesteix, A.L., Kneib, T., Augustin, T., Zeileis, A.: Conditional
  variable importance for random forests. BMC bioinformatics  \textbf{9}(1),
  ~307 (2008)

\bibitem{strobl2007bias}
Strobl, C., Boulesteix, A.L., Zeileis, A., Hothorn, T.: Bias in random forest
  variable importance measures: Illustrations, sources and a solution. BMC
  bioinformatics  \textbf{8}(1), ~25 (2007)

\bibitem{vstrumbelj2014explaining}
{\v{S}}trumbelj, E., Kononenko, I.: Explaining prediction models and individual
  predictions with feature contributions. Knowledge and information systems
  \textbf{41}(3),  647--665 (2014)

\bibitem{sundararajan2019many}
Sundararajan, M., Najmi, A.: The many {S}hapley values for model explanation.
  arXiv preprint arXiv:1908.08474  (2019)

\bibitem{sundararajan2017axiomatic}
Sundararajan, M., Taly, A., Yan, Q.: Axiomatic attribution for deep networks.
  arXiv preprint arXiv:1703.01365  (2017)

\bibitem{tibshirani1996regression}
Tibshirani, R.: Regression shrinkage and selection via the lasso. Journal of
  the Royal Statistical Society: Series B (Methodological)  \textbf{58}(1),
  267--288 (1996)

\bibitem{tolomei2017interpretable}
Tolomei, G., Silvestri, F., Haines, A., Lalmas, M.: Interpretable predictions
  of tree-based ensembles via actionable feature tweaking. In: Proceedings of
  the 23rd ACM SIGKDD international conference on knowledge discovery and data
  mining. pp. 465--474 (2017)

\bibitem{ustun2016supersparse}
Ustun, B., Rudin, C.: Supersparse linear integer models for optimized medical
  scoring systems. Machine Learning  \textbf{102}(3),  349--391 (2016)

\bibitem{ustun2019actionable}
Ustun, B., Spangher, A., Liu, Y.: Actionable recourse in linear classification.
  In: Proceedings of the Conference on Fairness, Accountability, and
  Transparency. pp. 10--19 (2019)

\bibitem{vapnik1974theory}
Vapnik, V., Chervonenkis, A.: Theory of pattern recognition (1974)

\bibitem{vilone2020explainable}
Vilone, G., Longo, L.: Explainable artificial intelligence: a systematic
  review. arXiv preprint arXiv:2006.00093  (2020)

\bibitem{visani2020optilime}
Visani, G., Bagli, E., Chesani, F.: Optilime: Optimized {LIME} explanations for
  diagnostic computer algorithms. arXiv preprint arXiv:2006.05714  (2020)

\bibitem{wachter2017counterfactual}
Wachter, S., Mittelstadt, B., Russell, C.: Counterfactual explanations without
  opening the black box: Automated decisions and the gdpr. Harv. JL \& Tech.
  \textbf{31}, ~841 (2017)

\bibitem{wang2015falling}
Wang, F., Rudin, C.: Falling rule lists. In: Artificial Intelligence and
  Statistics. pp. 1013--1022 (2015)

\bibitem{watson2019testing}
Watson, D.S., Wright, M.N.: Testing conditional independence in supervised
  learning algorithms. arXiv preprint arXiv:1901.09917  (2019)

\bibitem{wei2015variable}
Wei, P., Lu, Z., Song, J.: Variable importance analysis: a comprehensive
  review. Reliability Engineering \& System Safety  \textbf{142},  399--432
  (2015)

\bibitem{wexler2019if}
Wexler, J., Pushkarna, M., Bolukbasi, T., Wattenberg, M., Vi{\'e}gas, F.,
  Wilson, J.: The what-if tool: Interactive probing of machine learning models.
  IEEE transactions on visualization and computer graphics  \textbf{26}(1),
  56--65 (2019)

\bibitem{williamson2020efficient}
Williamson, B.D., Feng, J.: Efficient nonparametric statistical inference on
  population feature importance using {S}hapley values. arXiv preprint
  arXiv:2006.09481  (2020)

\bibitem{zeileis2008model}
Zeileis, A., Hothorn, T., Hornik, K.: Model-based recursive partitioning.
  Journal of Computational and Graphical Statistics  \textbf{17}(2),  492--514
  (2008)

\bibitem{zeiler2014visualizing}
Zeiler, M.D., Fergus, R.: Visualizing and understanding convolutional networks.
  In: European conference on computer vision. pp. 818--833. Springer (2014)

\bibitem{zhang2018interpretable}
Zhang, Q., Nian~Wu, Y., Zhu, S.C.: Interpretable convolutional neural networks.
  In: Proceedings of the IEEE Conference on Computer Vision and Pattern
  Recognition. pp. 8827--8836 (2018)

\bibitem{zhou2018measuring}
Zhou, Q., Liao, F., Mou, C., Wang, P.: Measuring interpretability for different
  types of machine learning models. In: Pacific-Asia Conference on Knowledge
  Discovery and Data Mining. pp. 295--308 (2018)

\bibitem{zou2005regularization}
Zou, H., Hastie, T.: Regularization and variable selection via the elastic net.
  Journal of the royal statistical society: series B (statistical methodology)
  \textbf{67}(2),  301--320 (2005)

\end{thebibliography}

\end{document}